# Residual Component Analysis: Generalising PCA for more flexible inference in linear-Gaussian models

Alfredo A. Kalaitzis                                   A.Kalaitzis@sheffield.ac.uk
Neil D. Lawrence                                       N.Lawrence@sheffield.ac.uk
Sheffield Institute for Translational Neuroscience and Department of Computer Science, University of Sheffield

## Abstract

Probabilistic principal component analysis (PPCA) seeks a low dimensional representation of a data set in the presence of independent spherical Gaussian noise. The maximum likelihood solution for the model is an eigenvalue problem on the sample covariance matrix. In this paper we consider the situation where the data variance is already partially explained by other factors, for example sparse conditional dependencies between the covariates, or temporal correlations leaving some residual variance. We decompose the residual variance into its components through a generalised eigenvalue problem, which we call residual component analysis (RCA). We explore a range of new algorithms that arise from the framework, including one that factorises the covariance of a Gaussian density into a low-rank and a sparse-inverse component. We illustrate the ideas on the recovery of a protein-signaling network, a gene expression time-series data set and the recovery of the human skeleton from motion capture 3-D cloud data.

## 1. Introduction

Probabilistic principal component analysis (PPCA) decomposes the covariance of a data vector $\mathbf{y}$ in $\mathbb{R}^p$, into a low-rank term and a spherical noise term. The underlying probabilistic model assumes that each datum is Gaussian distributed,

$$\mathbf{y} \sim \mathcal{N}(\mathbf{0}, \mathbf{W}\mathbf{W}^\top + \sigma^2 \mathbf{I}),$$

where we assume the data is centred such that their mean is zero and $\mathbf{W} \in \mathbb{R}^{p \times q}$, such that $q < p - 1$,

imposes a reduced rank structure on the covariance. The log-likelihood of the centered data set $\mathbf{Y}$ in $\mathbb{R}^{n \times p}$ with $n$ data points and $p$ features,

$$\ln p(\mathbf{Y}) = \sum_{i=1}^{n} \ln \mathcal{N}\left(\mathbf{y}_{i,:} \mid \mathbf{0},\ \mathbf{W}\mathbf{W}^\top + \sigma^2 \mathbf{I}\right),$$

can be maximized (Tipping & Bishop, 1999) with the result that $\mathbf{W}_{\text{ML}} = \mathbf{U}_q \mathbf{L}_q \mathbf{R}^\top$, where $\mathbf{U}_q$ are the $q$ principal eigenvectors of the sample covariance, $\tilde{\mathbf{S}} = n^{-1} \mathbf{Y}^\top \mathbf{Y}$, and $\mathbf{L}_q$ is a diagonal matrix with elements $\ell_{i,i} = (\lambda_i - \sigma^2)^{\frac{1}{2}}$, where $\lambda_i$ is the $i$th eigenvalue of the sample covariance and $\sigma^2$ is the noise variance. This maximum-likelihood solution is rotation invariant; that is, $\mathbf{R}$ is an arbitrary rotation matrix. As a result, the matrix $\mathbf{W}$ spans the principal subspace of the data and the model is known as *probabilistic*-PCA (PPCA). Underlying this model is an assumption that the data set is represented by $\mathbf{Y} = \mathbf{X}\mathbf{W}^\top + \mathbf{E}$, where $\mathbf{X}$ in $\mathbb{R}^{n \times q}$ is the matrix of $q$-dimensional latent variables and $\mathbf{E}$ is a matrix of noise variables $e_{i,j} \sim \mathcal{N}(0, \sigma^2)$. The marginal log-likelihood above is obtained by placing an isotropic prior independently on the elements of $\mathbf{X}$, $x_{i,j} \sim \mathcal{N}(0,1)$.

Lawrence (2005) showed that the PCA solution is also obtained for log-likelihoods of the form

$$\ln p(\mathbf{Y}) = \sum_{j=1}^{p} \ln \mathcal{N}(\mathbf{y}_{:,j} \mid \mathbf{0},\ \mathbf{X}\mathbf{X}^\top + \sigma^2 \mathbf{I})$$

which is recovered when we marginalize the loadings $\mathbf{W}$, instead of latent variables $\mathbf{X}$, with a Gaussian isotropic prior. This is the dual form of probabilistic PCA, also termed as *probabilistic principal coordinate analysis*, as this maximum likelihood solution solves for the latent coordinates $\mathbf{X}_{\text{ML}} = \mathbf{U}'_q \mathbf{L}_q \mathbf{R}^\top$, instead of the principal subspace basis. Here, $\mathbf{U}'_q$ are the first $q$ principal eigenvectors of the inner product matrix $p^{-1} \mathbf{Y}\mathbf{Y}^\top$ with $\mathbf{L}_q$ defined as before. Note that this Gaussian density is independent across *data features* rather than *data points*. So the correlation is expressed between data points. The underlying model is in fact a product of independent Gaussian processes (Rasmussen & Williams, 2006) with linear covariance functions.

Both primal and dual scenarios involve maximizing likelihoods of a similar covariance structure, namely when the covariance of the Gaussians is given by a low-rank term plus a spherical term, $\mathbf{XX}^\top + \sigma^2 \mathbf{I}$ (dual case). In this paper we consider a more general form of the covariance, given by $\mathbf{XX}^\top + \mathbf{\Sigma}$, where $\mathbf{\Sigma}$ is a *general positive definite* matrix. We are motivated by scenarios where our data has already been partly explained by the covariance matrix $\mathbf{\Sigma}$ and we wish to study the components of the residual variance. Our ideas can be applied in both primal and dual representations: the form to be used depends on what information we wish to encode in $\mathbf{\Sigma}$.

### 1.1. Motivating examples

Consider the general functional form of a *linear mixed-effects* model with two factors and noise, that will serve as a conceptual reference point for the rest of the paper (see Figure 1(a)):

$$\mathbf{Y} = \mathbf{XW}^\top + \mathbf{ZV}^\top + \mathbf{E}, \qquad (1)$$

where $\mathbf{Z}$ is a matrix of known covariates (*fixed* effects) with some predictive power for $\mathbf{Y}$, and $\mathbf{X}$ is a matrix of latent variables (*random* effects). The loadings $\mathbf{W}$ and $\mathbf{V}$ can be marginalised with Gaussian isotropic priors, to recover the log-likelihood

$$\ln p(\mathbf{Y}) = \sum_{j=1}^{p} \ln \mathcal{N}(\mathbf{y}_{:,j} | \mathbf{0}, \mathbf{XX}^\top + \mathbf{\Sigma}), \qquad (2)$$

where $\mathbf{\Sigma} = \mathbf{ZZ}^\top + \sigma^2 \mathbf{I}$ is positive-definite.

Specifically, one might have $\mathbf{Y}$ in eq. (1) manifested as:

**(a)** a set of protein activation signals under various external stimuli (heterogeneous data), with $\mathbf{V} = \mathbf{I}$ and $\mathbf{Z}$ as a set of covariates (equidimensional to $\mathbf{Y}$) sharing sparse conditional dependencies. Sparse dependencies are valuable for parsimonious modelling but might be confounded by latent effects $\mathbf{X}$ induced by the heterogeneous experimental conditions on the measurements $\mathbf{Y}$. This alludes to the instantiation $\mathbf{\Sigma}_{GL} = \mathbf{\Lambda}^{-1}$ for sparse $\mathbf{\Lambda}$, which recovers a *low-rank plus sparse-inverse* parameterisation of the covariance in eq. (2). A sparse precision amounts to a sparsely connected Gaussian Markov random field (GMRF) as the graphical model of $\mathbf{Z}$; that is, each $\mathbf{z}_{i,:}$ is distributed from $\mathcal{N}(\mathbf{0}, \mathbf{\Lambda}^{-1})$, where the precision matrix $\mathbf{\Lambda}$ is sparse (Lauritzen, 1996).

**(b)** a set of $n$ gene expression profiles, where each profile is a concatenated time-series of $p_1 + p_2$ timepoints, with $p_1$ timepoints sampled under control conditions and $p_2$ timepoints sampled under test conditions. In this scenario, the instantiation $\mathbf{\Sigma}_{Gram} = \mathbf{K} + \sigma^2 \mathbf{I}$, with $K_{ij} = k(\mathbf{z}_i, \mathbf{z}_j)$ as a general Gram matrix defined by some covariance function $k : \mathbb{R}^D \times \mathbb{R}^D \to \mathbb{R}$, could help express temporal correlations in a time-series dataset. This gets close to the common practice of explicitly subtracting the result of a simpler model from the data and then analyzing the residual separately.

**(c)** $\mathbf{Y}$ as a set of $n$ patients' gene expression measurements ($p$ genes), with $\mathbf{Z}$ as the genotype of each patient and $\mathbf{X}$ as the unobserved environmental effects (confounders), see Fusi et al. (2012).

In all of these cases, one useful task would be to analyse the components of the residual $\mathbf{XX}^\top$ for the dual case (or $\mathbf{WW}^\top$ for the primal case; for brevity we refer mostly to the dual case), given $\mathbf{\Sigma}$ or some estimate thereof. This begs the question: *Given $\mathbf{\Sigma}$, how can we solve for $\mathbf{X}$ (respectively $\mathbf{W}$)? And more importantly, for what instantiations of $\mathbf{\Sigma}$ can we formulate useful new algorithms for machine learning?*.

### 1.2. Proposed approach

The key theoretical result of this paper in Section 2, eq. (3) shows that the maximum-likelihood solution for $\mathbf{X}$ is simply based on a generalised eigenvalue problem (GEP) on the sample-covariance matrix. Hence, the low-rank term $\mathbf{XX}^\top$ can be optimized for general $\mathbf{\Sigma}$, with the only constraint being the positive-definiteness of $\mathbf{\Sigma}$. We call this data analysis approach *residual component analysis* (RCA).

Secondly, the RCA approach gives rise to a range of new algorithms suited for the aforementioned scenarios. For instance, for scenario **(a)** we propose an EM/RCA hybrid algorithm for estimating both the low-rank and sparse-inverse factors, see section 3. For scenario **(b)** we present a pure RCA treatment: the *residual* basis of interest is explored with a single exact estimate via the RCA GEP. We demonstrate the efficacy of the algorithms on biological and motion capture datasets in Section 4.

### 1.3. Related work and connection to CCA

The *low-rank plus sparse-inverse* parameterisation extends the *Graphical Lasso* algorithm (GLASSO, Friedman et al., 2008), which finds a MAP estimate of the covariance with an L1 regularization term on the precision. Sparse-inverse structures capture relations between variables that are not well characterized by low-rank forms. As such, the combination of sparse in-verse and low-rank can be a powerful one with applications in computational biology and visualisation, as we demonstrate in Section 4. We also point to the work of Stegle et al. (2011) for a different approach based on a multiplicative (Kronecker product) structure in the covariance.

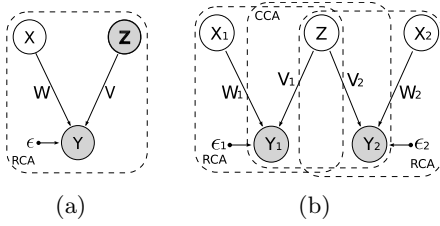

Figure 1. (a) A linear mixed-effects model. Fixed effects $\mathbf{Z}$ partially explain the variation in observations $\mathbf{Y}$. The *residual* covariance is spanned by $\mathbf{W}$ up to noise variance $\epsilon$. (b) Linear multi-view learning model where $\mathbf{Y}_1, \mathbf{Y}_2$ have shared ($\mathbf{Z}$) and private ($\mathbf{X}_1, \mathbf{X}_2$) latent components.

In addition, we note a few more connections to well-studied algorithms for linear dimensionality reduction: The obvious connection to PPCA is recovered by $\mathbf{\Sigma}_{pca} = \sigma^2 \mathbf{I}$. Furthermore, a *sparse*[1] $\mathbf{\Sigma}$ relates to the *robust*-PCA framework of Candes et al. (2009). Probabilistic CCA (Bach & Jordan, 2005) is recovered with $\mathbf{\Sigma}_{cca} = \begin{bmatrix} \mathbf{Y}_1\mathbf{Y}_1^\top & \mathbf{0} \\ \mathbf{0} & \mathbf{Y}_2\mathbf{Y}_2^\top \end{bmatrix} + \sigma^2 \mathbf{I}$ for $\mathbf{Y} = \begin{bmatrix} \mathbf{Y}_1 \\ \mathbf{Y}_2 \end{bmatrix}$ (for a proof, see supp. material). Furthermore, if $\mathbf{Y} = \begin{bmatrix} \mathbf{X}_1\mathbf{W}_1^\top + \mathbf{Z}\mathbf{V}_1^\top + \mathbf{E}_1 \\ \mathbf{X}_2\mathbf{W}_2^\top + \mathbf{Z}\mathbf{V}_2^\top + \mathbf{E}_2 \end{bmatrix}$, then in addition to the standard shared latent space of $\mathbf{Z}$ found in CCA, the partitions of $\mathbf{Y}$ have their own associated private latent spaces of $\mathbf{X}_1$ and $\mathbf{X}_2$, see Figure 1(b). This is in fact an instantiation of a general multi-view learning model (Ek et al., 2008), the linear case of which was more closely studied by Klami & Kaski (2008) who termed it *extended probabilistic-CCA*. To optimise this model, an *iterative* treatment of RCA can be used: solve for $\mathbf{V}$ on one step by setting $\mathbf{\Sigma}_{cca+} = \begin{bmatrix} \mathbf{W}_1\mathbf{W}_1^\top & \mathbf{0} \\ \mathbf{0} & \mathbf{W}_2\mathbf{W}_2^\top \end{bmatrix} + \sigma^2 \mathbf{I}$ in the GEP of the full data sample covariance; on the other step solve for each of $\mathbf{W}_k, k \in \{1, 2\}$, by setting $\mathbf{\Sigma}_{cca+} = \mathbf{V}_k\mathbf{V}_k^\top + \sigma^2 \mathbf{I}$ in the GEP of the sample covariance associated with $\mathbf{Y}_k$. This *iterative-RCA* algorithm is reminiscent of the expectation maximization (EM) algorithm for optimising *extended PCCA*, as both approaches maximize the likelihood by fitting components into the residual. More details on the CCA connection and *iterative-RCA* can be found in the supplementary material.

---

[1] As opposed to being an *inverse*-sparse matrix.

## 2. Maximum-Likelihood RCA

We show the main results on the dual case, with no loss of generalisation on the primal case.

**Theorem.** *The maximum likelihood estimate of the parameter $\mathbf{X}$ in the likelihood model in eq. (2), for positive-definite and invertible $\mathbf{\Sigma}$, is*

$$\mathbf{X}_{\mathrm{ML}} = \mathbf{\Sigma}\mathbf{S}(\mathbf{D} - \mathbf{I})^{\frac{1}{2}}, \tag{3}$$

*where $\mathbf{S}$ is the solution to the GEP*

$$\tfrac{1}{p}\mathbf{Y}\mathbf{Y}^\top \mathbf{S} = \mathbf{\Sigma}\mathbf{S}\mathbf{D}, \tag{4}$$

*with its columns as the generalised eigenvectors and $\mathbf{D}$ is diagonal with the corresponding generalised eigenvalues.*

**Proof.** The RCA log-likelihood is given by

$$L(\mathbf{X}, \mathbf{\Sigma}) = -\tfrac{p}{2}\ln|\mathbf{K}| - \tfrac{1}{2}\mathrm{tr}(\mathbf{Y}\mathbf{Y}^\top \mathbf{K}^{-1}) - \tfrac{np}{2}\ln(2\pi),$$

where $\mathbf{K} = \mathbf{X}\mathbf{X}^\top + \mathbf{\Sigma}$. Since $\mathbf{\Sigma}$ is positive-definite, we consider its eigen-decomposition $\mathbf{\Sigma} = \mathbf{U}\mathbf{\Lambda}\mathbf{U}^\top$, (5) where $\mathbf{U}^\top\mathbf{U} = \mathbf{I}$ and $\mathbf{\Lambda}$ is diagonal. The projection of the covariance onto this eigen-basis, scaled by the eigenvalues, is

$$\begin{aligned}\hat{\mathbf{K}} &= \mathbf{\Lambda}^{-\frac{1}{2}}\mathbf{U}^\top \mathbf{K}\mathbf{U}\mathbf{\Lambda}^{-\frac{1}{2}} \\ &= \mathbf{\Lambda}^{-\frac{1}{2}}\mathbf{U}^\top \mathbf{X}\mathbf{X}^\top \mathbf{U}\mathbf{\Lambda}^{-\frac{1}{2}} + \mathbf{I}.\end{aligned} \tag{6}$$

Letting $\hat{\mathbf{X}} = \mathbf{\Lambda}^{-\frac{1}{2}}\mathbf{U}^\top \mathbf{X}$ and $\hat{\mathbf{Y}} = \mathbf{\Lambda}^{-\frac{1}{2}}\mathbf{U}^\top \mathbf{Y}$, allows us to define $\hat{\mathbf{K}} = \hat{\mathbf{X}}\hat{\mathbf{X}}^\top + \mathbf{I}$ and its inverse

$$\hat{\mathbf{K}}^{-1} = \mathbf{\Lambda}^{\frac{1}{2}}\mathbf{U}^\top \mathbf{K}^{-1}\mathbf{U}\mathbf{\Lambda}^{\frac{1}{2}}. \tag{7}$$

Therefore from (7), we have $|\mathbf{K}| = |\hat{\mathbf{K}}||\mathbf{\Lambda}|$ and $\mathrm{tr}(\mathbf{Y}\mathbf{Y}^\top \mathbf{K}^{-1}) = \mathrm{tr}(\mathbf{\Lambda}^{-\frac{1}{2}}\mathbf{U}^\top \mathbf{Y}\mathbf{Y}^\top \mathbf{U}\mathbf{\Lambda}^{-\frac{1}{2}}\hat{\mathbf{K}}^{-1})$. Now we can rewrite the entire RCA log-likelihood in terms of the transformed variables $\hat{\mathbf{X}}$ and $\hat{\mathbf{Y}}$,

$$L(\hat{\mathbf{X}}) = -\tfrac{p}{2}\ln(|\hat{\mathbf{K}}||\mathbf{\Lambda}|) - \tfrac{1}{2}\mathrm{tr}(\hat{\mathbf{Y}}\hat{\mathbf{Y}}^\top \hat{\mathbf{K}}^{-1}) - \tfrac{np}{2}\ln(2\pi).$$

We know how to maximize this new form of the log-likelihood with respect to $\hat{\mathbf{X}}$. Following a similar route to the maximum likelihood solution proof in Tipping & Bishop (1999), the gradient

$$\frac{\partial L}{\partial \hat{\mathbf{X}}} = \hat{\mathbf{K}}^{-1}\hat{\mathbf{Y}}\hat{\mathbf{Y}}^\top \hat{\mathbf{K}}^{-1}\hat{\mathbf{X}} - p\hat{\mathbf{K}}^{-1}\hat{\mathbf{X}}, \tag{8}$$

gives the stationary point $\tfrac{1}{p}\hat{\mathbf{Y}}\hat{\mathbf{Y}}^\top \hat{\mathbf{K}}^{-1}\hat{\mathbf{X}} = \hat{\mathbf{X}}$. (9)

By substituting the singular value decomposition

$$\hat{\mathbf{X}} = \hat{\mathbf{V}}\mathbf{L}\mathbf{R}^\top, \quad \text{for} \quad \hat{\mathbf{X}} \quad \text{in (8),} \quad \text{gives} \tag{10}$$

$\hat{\mathbf{V}}\mathbf{L}\mathbf{R}^\top = \frac{1}{p}\hat{\mathbf{Y}}\hat{\mathbf{Y}}^\top(\hat{\mathbf{V}}\mathbf{L}^2\hat{\mathbf{V}}^\top + \mathbf{I})^{-1}\hat{\mathbf{V}}\mathbf{L}\mathbf{R}^\top$. By the *Woodbury matrix identity*, we can see that maximisation relies on the regular eigenvalue problem

$$\frac{1}{p}\hat{\mathbf{Y}}\hat{\mathbf{Y}}^\top\hat{\mathbf{V}} = \hat{\mathbf{V}}\mathbf{D}, \quad \text{where} \quad \mathbf{D} = (\mathbf{L}^2 + \mathbf{I}). \quad (11)$$

Next we focus on relating the stationary point of $\hat{\mathbf{X}}$ to the solution for $\mathbf{X}$ and then we proceed by expressing this eigenvalue problem in terms of $\mathbf{Y}\mathbf{Y}^\top$: By the definition of $\hat{\mathbf{X}}$, we obtain the factorisation

$$\mathbf{X} = \mathbf{U}\mathbf{\Lambda}^{\frac{1}{2}}\hat{\mathbf{V}}\mathbf{L}\mathbf{R}^\top = \mathbf{T}\mathbf{L}\mathbf{R}^\top, \quad (12)$$

where we have defined $\mathbf{T} = \mathbf{U}\mathbf{\Lambda}^{\frac{1}{2}}\hat{\mathbf{V}}$. Since $\mathbf{\Sigma}$ is invertible, we substitute for $\hat{\mathbf{Y}}$ and $\hat{\mathbf{V}}$ in (11) and use the inverse of (5) to recover the equivalent eigenvalue problem in the original dual-space

$$\frac{1}{p}\mathbf{Y}\mathbf{Y}^\top\mathbf{\Sigma}^{-1}\mathbf{T} = \mathbf{T}\mathbf{D}.$$

To conclude the proof, we define $\mathbf{S} = \mathbf{\Sigma}^{-1}\mathbf{T}$ to recover the desired *symmetric* form of the GEP

$$\frac{1}{p}\mathbf{Y}\mathbf{Y}^\top\mathbf{S} = \mathbf{\Sigma}\mathbf{S}\mathbf{D}.$$

Based on the factorisation of $\mathbf{X}$ in (12), now we can recover $\mathbf{X}$ up to an arbitrary rotation ($\mathbf{R}$, which for convenience is normally set to $\mathbf{I}$), via the first $q$ generalised eigenvectors of $p^{-1}\mathbf{Y}\mathbf{Y}^\top$,

$$\mathbf{X} = \mathbf{T}\mathbf{L} = \mathbf{\Sigma}\mathbf{S}\mathbf{L} = \mathbf{\Sigma}\mathbf{S}(\mathbf{D} - \mathbf{I})^{\frac{1}{2}}. \quad \square$$

**Commentary.** Due to the algebraic symmetry between the dual and primal formulations of the log-marginal likelihood in (2), one can easily extend the proof to the primal case. Specifically, the maximum likelihood solution of $\mathbf{W}$ in (1) has the same form $\mathbf{W} = \mathbf{\Sigma}\mathbf{S}(\mathbf{D} - \mathbf{I})^{\frac{1}{2}}$, where now $\frac{1}{n}\mathbf{Y}^\top\mathbf{Y}\mathbf{S} = \mathbf{\Sigma}\mathbf{S}\mathbf{D}$ and $\mathbf{\Sigma} = \mathbf{V}\mathbf{V}^\top + \sigma^2\mathbf{I}$.

Aside from $\mathbf{\Sigma}$, we note a subtle difference from the PPCA solution for $\mathbf{W}$: Whereas PPCA explicitly subtracts the noise variance from the $q$ retained principal eigenvalues, RCA in (6) implicitly incorporates any noise terms into $\mathbf{\Sigma}$ and *standardises* them when it projects the total covariance onto the eigen-basis of $\mathbf{\Sigma}$. Thus we get a reduction of unity from the retained generalised eigenvalues in (3). Again, for $\mathbf{\Sigma} = \mathbf{I}$ the two solutions are identical.

Finally, we state the posterior density for the RCA probabilistic model (primal case) and $\boldsymbol{\mu}_y = \mathbf{0}$,

$$\begin{aligned}\mathbf{x}|\mathbf{y} &\sim \mathcal{N}(\mathbf{\Sigma}_{\mathbf{x}|\mathbf{y}}\mathbf{W}_{\mathrm{ML}}^\top\mathbf{\Sigma}^{-1}\mathbf{y}\,,\,\mathbf{\Sigma}_{\mathbf{x}|\mathbf{y}}),\\ \text{where}\quad \mathbf{\Sigma}_{\mathbf{x}|\mathbf{y}} &= (\mathbf{W}_{\mathrm{ML}}^\top\mathbf{\Sigma}^{-1}\mathbf{W}_{\mathrm{ML}} + \mathbf{I})^{-1}.\end{aligned} \quad (13)$$

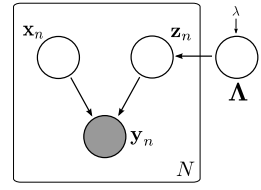

Figure 2. Graphical model optimised by the EM/RCA hybrid algorithm.

## 3. Low Rank Plus Sparse Inverse

In this section we show how to optimise the following generative model, summarised in Figure 2,

$$\begin{aligned}\mathbf{y}|\mathbf{x}, \mathbf{z} &\sim \mathcal{N}(\mathbf{W}\mathbf{x} + \mathbf{z}, \sigma^2\mathbf{I}),\\ \mathbf{x} &\sim \mathcal{N}(\mathbf{0}, \mathbf{I}), \quad \mathbf{z} \sim \mathcal{N}(\mathbf{0}, \mathbf{\Lambda}^{-1}),\end{aligned} \quad (14)$$

where $\mathbf{\Lambda}$ is sampled from a Laplace prior density,

$$p(\mathbf{\Lambda}) \propto \exp(-\lambda\|\mathbf{\Lambda}\|_1).$$

Marginalising $\mathbf{X}$, yields
$\log p(\mathbf{Y}, \mathbf{\Lambda}) = \sum_{i=1}^n \log\{\mathcal{N}(\mathbf{y}_{i,:}|\mathbf{0}, \mathbf{W}\mathbf{W}^\top + \mathbf{\Sigma}_{GL})p(\mathbf{\Lambda})\}$

$$\geq \int q(\mathbf{Z})\log\frac{p(\mathbf{Y}, \mathbf{Z}, \mathbf{\Lambda})}{q(\mathbf{Z})}\mathrm{d}\mathbf{Z} \quad (15)$$

where $q(\mathbf{Z})$ is the variational distribution and $\mathbf{\Sigma} = \mathbf{\Lambda}^{-1} + \sigma^2\mathbf{I}$, which we wish to optimise for some known $\mathbf{W}$. This is an intractable problem, so instead we optimise the lower bound (15) in an EM fashion.

**E-step** Replacing $q(\mathbf{Z})$ with the posterior $p(\mathbf{Z}|\mathbf{Y}, \mathbf{\Lambda}')$ for a current estimate $\mathbf{\Lambda}'$, amounts to the E-step for updating the posterior density of $\mathbf{z}_n|\mathbf{y}_n$ with

$$\mathrm{cov}[\mathbf{z}|\mathbf{y}] = ((\mathbf{W}\mathbf{W}^\top + \sigma^2\mathbf{I})^{-1} + \mathbf{\Lambda}')^{-1} \quad (16)$$
$$\langle\mathbf{z}_n|\mathbf{y}_n\rangle = \mathrm{cov}[\mathbf{z}_n|\mathbf{y}_n](\mathbf{W}\mathbf{W}^\top + \sigma^2)^{-1}\mathbf{y}_n \quad (17)$$
$$\text{and}\quad \langle\mathbf{z}_n\mathbf{z}_n^\top\rangle = \mathrm{cov}[\mathbf{z}|\mathbf{y}] + \langle\mathbf{z}_n\rangle\langle\mathbf{z}_n\rangle^\top. \quad (18)$$

**M-step** Then for fixed $\mathbf{Z}'$, the only free parameter in the expected complete-data log-likelihood $\mathcal{Q} = \mathbb{E}_{\mathbf{Z}|\mathbf{Y}}(\log p(\mathbf{Z}', \mathbf{\Lambda}))$ is $\mathbf{\Lambda}$. Therefore, $\operatorname*{argmax}_{\mathbf{\Lambda}}\mathcal{Q} =$

$$\operatorname*{argmax}_{\mathbf{\Lambda}}\left(\frac{n}{2}\ln|\mathbf{\Lambda}| - \frac{1}{2}\sum_{i=1}^n \mathrm{tr}(\langle\mathbf{z}_n\mathbf{z}_n^\top\rangle\mathbf{\Lambda})) - \frac{n}{2}\lambda\|\mathbf{\Lambda}\|_1\right), \quad (19)$$

which amounts to standard GLASSO optimisation with the covariance matrix from (18).

**RCA-step** After one iteration of EM, we update $\mathbf{W}$ via RCA based on the newly estimated $\mathbf{\Lambda}$,

$$\begin{aligned}\mathbf{W} &= \mathbf{\Sigma}\mathbf{S}(\mathbf{D} - \mathbf{I})^{\frac{1}{2}}, \quad \text{for the GEP}\\ \frac{1}{n}\mathbf{Y}^\top\mathbf{Y}\mathbf{S} &= \mathbf{\Sigma}\mathbf{S}\mathbf{D} \quad \text{and} \quad \mathbf{\Sigma} = \mathbf{\Lambda}^{-1} + \sigma^2\mathbf{I}.\end{aligned} \quad (20)$$

Algorithm (1) summarises the EM and RCA steps which is collectively one iteration of EM/RCA:

---

**Algorithm 1** EM/RCA

Initialise $\sigma^2$, $\mathbf{W}$ and $\mathbf{\Lambda}$.
**repeat**
    **E-step:** Update posterior density of $\mathbf{Z}|\mathbf{Y}$ with (16) and (17).
    **M-step:** Update $\mathbf{\Lambda}$ with (19).
    **RCA-step:** Update $\mathbf{W}$ with (20).
**until** the lower-bound (15) converges

---

## 4. Experiments

We describe three experiments with EM/RCA and one purely with RCA analysing the residual left from a Gaussian process (GP) in a time-series. For all the experiments that involve EM/RCA, the following apply:

Initialisations are $\sigma^2 = \frac{1}{2p}\text{tr}(\mathbf{C}_y)$, where $\mathbf{C}_y$ is the sample covariance of the analysed data; $\mathbf{W} = \mathbf{U}_q(\mathbf{L}_q - \sigma^2\mathbf{I})^{\frac{1}{2}}$ as the q principal eigenvectors whose eigenvalues are larger than $\sigma^2$, and $\mathbf{\Lambda} = \mathbf{I}$. Note that since $\sigma^2$ is fixed to its initialized value, this implicitly fixes the number of latent variables. A more systematic approach would be a line search on $\sigma^2$ during the M-step or using the BIC criterion over a small range of $q$ (number of latent variables).

Each dataset was analysed with $\ell_1$-regularisation parameters $\lambda = 5^x$ for $x$ linearly interpolated in the interval $[-8, 3]$, thus creating a solution path as $\lambda$ increases exponentially. For lasso-based algorithms, in general, the solution paths tend to be unstable, so to smoothen the solution paths we applied stability selection (Meinshausen & Bühlmann, 2010), i.e. for each dataset and for each $\lambda$, results of all methods (GLASSO, Kronecker-GLASSO, EM/RCA) are stabilized by taking 100 repeats using 90% of the datapoints for each repeat. Edges (corresponding to connections in the figure) are assumed to be present if they are called more than 50% of the time.

### 4.1. Simulation

First we consider an artificial dataset sampled from the generative model in (equation 14, Figure 2) to illustrate the effects of confounders on the estimation of the sparse-inverse covariance. Specifically,
$\mathbf{Y} = \mathbf{X}\mathbf{W}^\top + \mathbf{Z} + \mathbf{E}$, where $\mathbf{Y} \in \mathbb{R}^{100 \times 50}$; $\mathbf{W} \in \mathbb{R}^{50 \times 3}$; $\mathbf{X} \in \mathbb{R}^{100 \times 3}$, for each $\mathbf{x}_{i,:} \overset{\text{iid}}{\sim} \mathcal{N}(\mathbf{0}, \mathbf{I}_3)$; $\mathbf{z}_{i,:} \overset{\text{iid}}{\sim} \mathcal{N}(\mathbf{0}, \mathbf{\Lambda}^{-1})$; $\mathbf{\Lambda}$ was generated with a sparsity level of 1% over all possible edges in the GRF and its non-zero entries were iid-sampled from a Gaussian with mean

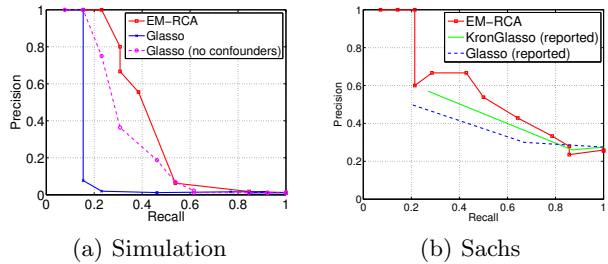

*Figure 3.* (a) Recall-precision curves of EM/RCA and GLASSO on simulated confounded data (solid line), and GLASSO on simulated non-confounded data (dashed line). (b) Similarly on Sachs data. Kronecker-GLASSO and GLASSO curves are identical to the ones reported in (Stegle et al., 2011).

1 and variance 2; $\mathbf{z}_{:,j} \overset{\text{iid}}{\sim} \mathcal{N}(\mathbf{0}, \gamma \mathbf{I}_{50})$; Finally for the noise, $\mathbf{e}_{i,:} \overset{\text{iid}}{\sim} \mathcal{N}(\mathbf{0}, \sigma^2 \mathbf{I})$. The variance $\gamma$ was set such that $\mathbf{\Lambda}^{-1}$ and $\mathbf{W}\mathbf{W}^\top$ explained equal variance and $\sigma^2$ was set such that the signal-to-noise ratio was 10.

Figure 3(a) shows the precision-recall curve for GLASSO and EM/RCA. The EM/RCA curve shows significantly better performance than GLASSO on the confounded data, while the dashed line shows the performance of GLASSO on similarly generated data without the confounding effects ($\mathbf{W} = \mathbf{0}$). We note that EM/RCA performs better on confounded data than GLASSO on non-confounded data, because of the lower signal-to-noise ratio in the non-confounded data.

### 4.2. Reconstruction of a biological network

Next we applied EM/RCA on the protein-signaling data of Sachs et al. (2008). In this case, we also compare against the results reported by Stegle et al. (2011) on the same data, with the *Kronecker-GLASSO* algorithm. These data provide signal measurements from 11 proteins under various external stimuli. We combined measurement from the first 3 experiments, resulting in a heterogeneous dataset of 2,666 samples. The different conditions of these experiments induce the confounding effects in the data. For the sake of comparison, we also run the analysis on a random 10% subset of the 2,666 samples. All algorithm were compared based on the moralised-version of the ground truth directed network which was biologically validated in the related study.

In Figure 3(b), EM/RCA performs slightly better than all other methods. Figure 4 shows the reconstructed networks for recall 0.4. We note that EM/RCA is more conservative in calling edges.

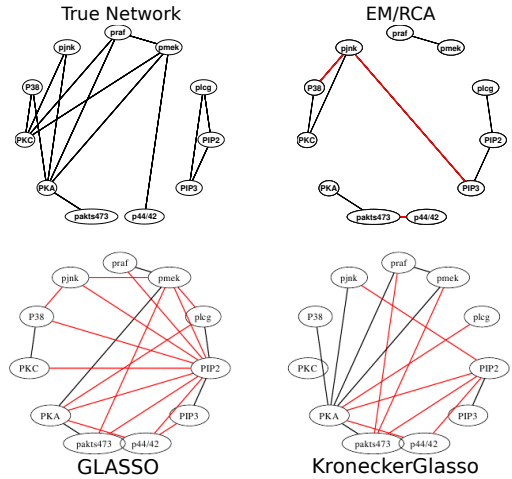

Figure 4. Networks reconstructed (recall 0.4) via EM/RCA, Kronecker-GLASSO and GLASSO on Sachs data. Red edges signify false positives.

### 4.3. Reconstruction of the human form

These data come from the CMU motion-capture database (http://mocap.cs.cmu.edu). The objective here is to reconstruct the underlying connectivity of a human being, given only the 3 dimensional locations of 31 sensors placed about the figures body. Each captured motion in the database involves the skeleton (or stickman) data specific to the person under the trial (different heights, builds, etc.) and the 3-D sensor cloud data. Each trial involves 31 sensors, so the dataset for each trial is 93 × the number of frames captured in the trial.

Our aim is to construct a model which recovers connectivity between these points. This should be possible because we expect sensors that are connected in the underlying figure to be conditionally independent of other sensors in the figure. This motivates the underlying sparse structure. Conversely, different motions exhibit much broader correlations across the figure. In particular walking exhibits anti-correlations between sensors on different legs and across the arms. These types of motion should be far better recovered through a low rank representation of the covariance.

If, as expected, the raw data is confounded by low-rank properties associated with particular structured motions (as opposed to random poses, as might be adopted by a wooden artist's doll) then our combination of low-rank with sparse connectivity should outperform a model based purely on sparse connectivity. We therefore compare EM/RCA and GLASSO on trials involving walking, running, jumping and dancing. The local connectivity between the sensors, i.e. the human skeleton, should be represented in the sparse matrix $\mathbf{\Lambda}$ (proscribing a Gaussian random field). To further motivate this idea we note that $\mathbf{\Lambda}$ can also be seen as the stiffness (or Laplacian) matrix of a physical system of a spring network, where the off-diagonal entries represent the negative stiffness of the spring. To detect a connection between two sensors we are only looking for negative entries in the estimated $\mathbf{\Lambda}$.

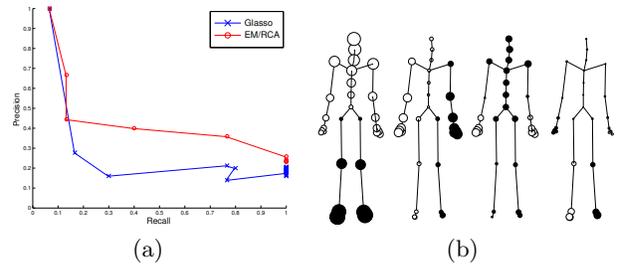

Figure 5. (a) Recall-precision curves of EM/RCA and GLASSO on the CMU motion-capture data. (b) Hinton diagram of $\mathbf{X}$ capturing the confounding effects in the motions. Each column of $\mathbf{X}$ is visualised by rearranging its elements to the corresponding sensors on the ground truth stickman. The colour of a dot indicates the sign and the size is proportional to the magnitude of the corresponding element in $\mathbf{X}$.

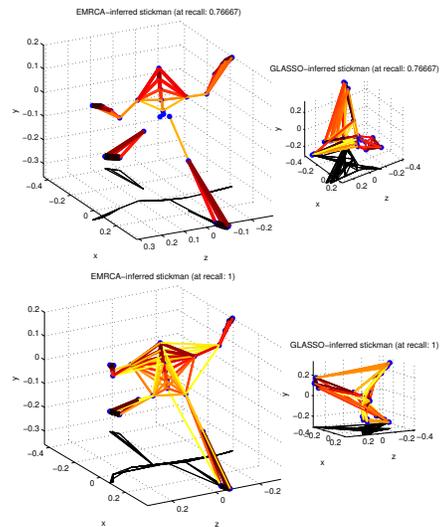

Figure 6. Network reconstructions via EM/RCA (left) and GLASSO (right) for recall 0.77 (top) and 1 (bottom). Inferred edges are superimposed on the *characteristic poses* extracted from the 3 principal eigenvectors of the estimated sparse $\mathbf{\Lambda}$ (or Laplacian of the spring system). Edge color indicates the negative stiffness intensity (red is large) and the black lines are shadows for aiding the perspective.

Figure 5 shows the results in the form of recall/precision curves for both GLASSO and our EM/RCA implementation of a sparse-inverse plus low-rank model. The EM/RCA algorithm consistently outperforms standard GLASSO. Figure 6 illustrates the recovered stickmen of EM/RCA and GLASSO. We note that the connectivities and eigenposes are more faithful to the true human form, in comparison to GLASSO. For a small $\lambda$ setting (recall 1) the precisions are similar; nonetheless the human form is robust, with very weak (yellow) edges wherever they do not apply (e.g. elbow-waist, elbow-head). This signifies that the precision measure might be ill-suited for evaluating a stickman, where the network configuration has a spatial interpretation. The ground truth is also "noisy" in the sense that a shoulder-chest edge, for instance, must be called as the torso is a rigid part of the human body (high stiffness). Figure 5(b) illustrates the confounding effects that $\mathbf{X}$ is supposed to capture. Specifically, in the first component, the legs are anti-correlated to the upper-half of the body, which can be attributed to jumping motions. The second and forth components capture anti-correlations across the different legs and arms, exhibited by walking and running, as discussed earlier. The third component shows strong anti-correlation between the hands and the rest of the upper-body, which is more open to interpretation.

### 4.4. Differences in gene-expression profiles

A common challenge in data analysis is to summarize the difference between treatment and control samples. To illustrate how RCA can help, we consider two gene expression time-series of cell lines. The treatment cells are targeted by TP63 introduced into the nucleus by tamoxifen. The control cells are simply subject to tamoxifen alone. The data used for this case study come from Della Gatta et al. (2008, GEO accession GSE10562) The treatment group $\mathbf{Y}_1 \in \mathbb{R}^{n_1 \times p}$ contains n=13 time-points of $p = 22,690$ gene expression measurements, whilst the control group $\mathbf{Y}_2 \in \mathbb{R}^{n_2 \times p}$ contains only $n_2 = 7$ time-points. This complexity of data (with different numbers of time-points and non-uniform sampling) is typical of many bio-medical data sets. The challenge is to represent the differences between the gene expression profiles for these two data sets. CCA could be applied but this would represent the similarities between the data, not the differences.

Assuming that both time-series are identical, implies $\mathbf{y}^\top = (\mathbf{y}_1^\top\ \mathbf{y}_2^\top)$ can be modeled by a Gaussian process (GP) with a temporal covariance function, $\mathbf{y} \sim \mathcal{N}(\mathbf{0}, \mathbf{K})$, where $\mathbf{K} \in \mathbb{R}^{n \times n}$ for $n=n_1+n_2$ is structured such that both $\mathbf{y}_1$ and $\mathbf{y}_2$ are generated from the same function, $K_{i,j} = k(t_i, t_j) = \exp(-\frac{1}{2}\ell^{-2}(t_i - t_j)^2)$, a squared-exponential covariance function (or RBF ker-nel, Figure 7(a)). Other kernels can be used and their hyperparameters can be optimized, but for this demonstration we simply set $\ell = 20$ which provides a bandwidth roughly in line with the time-point sampling intervals. We also add a small noise term along the diagonal of $\mathbf{K}$ which was set to 1% of the data variance.

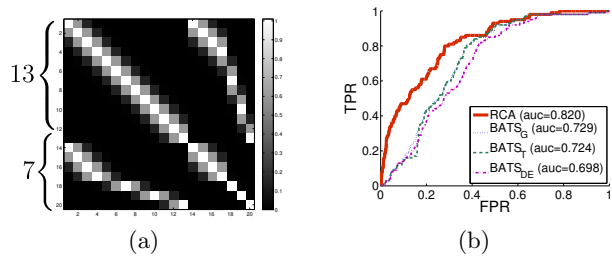

Figure 7. (a) RBF kernel computed on augmented time-input vectors of gene-expression. The kernel is computed across times $(0:20:240, 0, 20, 40, 60, 120, 180, 240)$, jointly for control and treatment. (b) ROC curves of RCA and BATS variants with different noise models (G: Gaussian, T: t-distribution, DE: double-exponential). See also (Kalaitzis & Lawrence, 2011; Stegle et al., 2010) GP-based approaches.

A more general model (dual paradigm) of the form $\mathbf{y} \sim \mathcal{N}(\mathbf{0}, \mathbf{X}\mathbf{X}^\top + \mathbf{K})$, should explain no variance in the low-rank component $\mathbf{X}\mathbf{X}^\top$, as all the signal in the time-series is assumed to be explained by the underlying function sampled from the GP. If we solve for the residual components $\mathbf{X}$ via RCA, they will be forced to explain how the two time-series are actually different.

We project the profiles onto the eigenbasis of the first $q$ generalised eigenvectors $\tilde{\mathbf{Y}} = \mathbf{S}_q^\top \mathbf{Y}$ and obtain a score of differential expression based on the norms of their projections. The number $q$ of retained principal eigenvectors is decided on the number of corresponding eigenvalues $d_i$ being larger than one. Recall from PPCA, that as we increase the assumed noise variance $\sigma^2$, more eigenvalues become negative and less eigenvectors are retained in $\mathbf{W}_{\mathrm{ML}}$ (cf. page 1). Similarly, RCA standardises any noise (6), so we only have to retain for eigenvalues larger than 1. In this case, the assumed noise variance embedded in the kernel drives the effective number of eigenvectors in the projection basis. Ranking the scores and comparing to the noisy ground-truth list of binding targets[2] of TP63 from (Della Gatta et al., 2008), gives the ROC performance curve in Figure 7(b). We compare against BATS as a baseline method (Angelini et al., 2007). We note that RCA outperforms BATS in terms the area under the ROC curve for all of its noise models.

---

[2] A gene with a large number of binding sites for TP63 is a strong candidate for being one of its direct targets, and thus associated with TP63-related diseases.

## 5. Discussion

We are often faced with data that can be partially explained by a set of covariates and may wish to analyse the residual components of these data. This motivated the construction of RCA: an algorithm for describing a low-dimensional representation of the residuals of a data set, given partial explanation by a covariance matrix $\Sigma$. The low-rank component of the model can be determined through a generalized eigenvalue problem. The special case of PCA being recovered for $\Sigma = \sigma^2 \mathbf{I}$. Our algorithm also generalizes CCA, but with further imaginative application we can develop new approaches to data analysis.

We illustrated how a treatment and a control time-series could have their differences highlighted through appropriate selection of $\Sigma$ (in this case we used an RBF kernel). We also introduced an algorithm for fitting a variant of CCA where the private spaces are explained through low dimensional latent variables.

Our final, and perhaps most important, new data analysis technique combined sparse-inverse covariance with low-rank. Full covariance matrix models of data are often problematic as their parameterization scales with $D^2$. Two separate approaches to a reduced parameterization of these matrices are to base them on low-rank matrices (as in probabilistic PCA) or on a sparse-inverse structure (as in GLASSO). These two approaches have very different characteristics: one involves specifying sparse conditional dependencies in the data, the other assumes that a reduced set of latent variables governs the data. Clearly, in any data set, both of these characteristics may be present. Our sparse-inverse plus low-rank approach is the first approach to deal with both of these cases in the same model. It was demonstrated to good effect in a motion capture and protein network example.